\documentclass[letterpaper]{article} 
\usepackage{aaai25}  
\usepackage{times}  
\usepackage{helvet}  
\usepackage{courier}  
\usepackage[hyphens]{url}  
\usepackage{graphicx} 
\usepackage{natbib}  
\usepackage{caption} 
\usepackage{algorithm}
\usepackage{algorithmic}
\usepackage{soul}
\usepackage{subfigure}
\usepackage{multirow}
\usepackage{booktabs}
\usepackage{amsthm,amsmath,amssymb}
\usepackage{mathrsfs}
\usepackage{arydshln}

\usepackage[inline]{enumitem}

\usepackage[symbol]{footmisc}
\usepackage{caption}

\usepackage{graphicx}
\usepackage{placeins}
\usepackage{amssymb}  
\usepackage{dsfont}

\DeclareMathOperator*{\argmax}{arg\,max}
\usepackage{newfloat}
\usepackage{listings}
\DeclareCaptionStyle{ruled}{labelfont=normalfont,labelsep=colon,strut=off} 
\floatstyle{ruled}
\newfloat{listing}{tb}{lst}{}
\floatname{listing}{Listing}
\title{Context-aware Inductive Knowledge Graph Completion with Latent Type Constraints and Subgraph Reasoning}
\author {
    Muzhi Li\textsuperscript{\rm 1,2}\thanks{Equal contribution.},
    Cehao Yang\textsuperscript{\rm 3,2}\footnotemark[1],
    Chengjin Xu\textsuperscript{\rm 2}\footnotemark[1],
    Zixing Song\textsuperscript{\rm 4},
    Xuhui Jiang\textsuperscript{\rm 2},
    Jian Guo\textsuperscript{\rm 2}\thanks{Corrsponding authors. }, \\
    Ho-fung Leung\thanks{Independent researcher.}, 
    Irwin King\textsuperscript{\rm 1}\footnotemark[2]
}
\affiliations {
    \textsuperscript{\rm 1}Department of Computer Science and Engineering, The Chinese University of Hong Kong\\
    \textsuperscript{\rm 2}IDEA Research, International Digital Economy Academy\\
    \textsuperscript{\rm 3}Artificial Intelligence Thrust, Hong Kong University of Science and Technology (Guangzhou)\\
    \textsuperscript{\rm 4}Department of Engineering, University of Cambridge 
    \\
    \{mzli,king\}@cse.cuhk.edu.hk, \{limuzhi,yangcehao,xuchengjin,guojian\}@idea.edu.cn, \\
    cyang289@connect.hkust-gz.edu.cn, zs456@cam.ac.uk, ho-fung.leung@outlook.com
}

\usepackage{bibentry}

\begin{document}
\maketitle
\begin{abstract}
Inductive knowledge graph completion~(KGC) aims to predict missing triples with unseen entities. 
Recent works focus on modeling reasoning paths between the head and tail entity as direct supporting evidence. However, these methods depend heavily on the existence and quality of reasoning paths, which limits their general applicability in different scenarios. In addition, we observe that latent type constraints and neighboring facts inherent in KGs are also vital in inferring missing triples. To effectively utilize all useful information in KGs, we introduce CATS, a novel context-aware inductive KGC solution. With sufficient guidance from proper prompts and supervised fine-tuning, CATS activates the strong semantic understanding and reasoning capabilities of large language models to assess the existence of query triples, which consist of two modules. First, the type-aware reasoning module evaluates whether the candidate entity matches the latent entity type as required by the query relation. Then, the subgraph reasoning module selects relevant reasoning paths and neighboring facts, and evaluates their correlation to the query triple. Experiment results on three widely used datasets demonstrate that CATS significantly outperforms state-of-the-art methods in 16 out of 18 transductive, inductive, and few-shot settings with an average absolute MRR improvement of $7.2\%$. 
\end{abstract}

%
\begin{links}
    \link{Code}{https://github.com/IDEA-FinAI/CATS}
\end{links}

\section{Introduction}
Knowledge Graphs~(KGs) are graph-structured knowledge bases that represent facts with triples in the form of (\textit{head entity, relation, tail entity}). KGs become essential in various downstream applications such as question answering~\cite{ToG}, fact checking~\cite{FactKG}, and recommendation systems~\cite{recommendation}. In practice, most real-world KGs are incomplete, which highlights the significance of the knowledge graph completion~(KGC) (or \textit{relation prediction}) task, which aims to predict the missing head or tail entity from the query triples.

Existing approaches to the KGC task usually perform well under an ``\textbf{\textit{transductive setting}}'', where missing entities can be observed in training triples. Conversely, the ``\textbf{\textit{inductive}}'' KGC task requires the model to handle newly emerged entities, which is more reflective of real-world scenarios, as KGs are continuously evolving. The inductive setting highlights the importance of three key contexts inherent in KGs, namely entity types, reasoning paths, and neighboring facts. 

\begin{figure}
    \centering
    \includegraphics[width=0.47\textwidth]{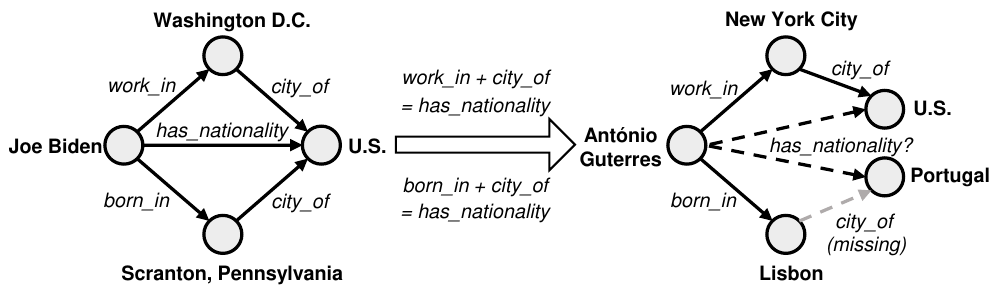}
    \caption{A typical scenario of inductive knowledge graph completion. The model needs to infer the nationality of the unseen entity ``António Guterres''. }
    \label{figure_intro}
\end{figure}

First, relations within KGs impose \textit{latent type constraints} to head and tail entities being connected, which are crucial in inferring potential missing triples.
For instance, the relation \textit{works in} typically connects a \textit{person} (head) and a \textit{location} (tail). Although we have not encountered newly emerged entities during training, a triple can still be considered plausible if the head and tail entities conform to the implicit types as required by the relation. 

Second, \textit{reasoning paths} provide direct clues for the existence of missing triple~\cite{BERTRL}. However, these paths can be unreliable in certain contexts. For example, in Figure~\ref{figure_intro}, if the training set contains numerous triples such as (\textit{Joe Biden}, \textit{works in}, \textit{Washington, D.C.}), (\textit{Washington, D.C.}, \textit{city of}, \textit{U.S.}), and (\textit{Joe Biden}, \textit{has nationality}, \textit{U.S.}), the model may come up with some rules like (works in + city of = has nationality). It should be noted that such implicit rules do not invariably apply. A notable counterexample is \textit{António Guterres}, who works in \textit{New York City} as the Secretary-General of the United Nations. 

Finally, \textit{neighboring facts} of the head and tail entities also provide valuable clues for triple completion.
For instance, in Figure~\ref{figure_intro}, it is difficult to predict entity ``\textit{António Guterres}'' has a Portuguese nationality solely based on the available reasoning path. Fortunately, the presence of specific neighboring facts such as (\textit{António Guterres}, \textit{born in}, \textit{Lisbon}) can help disambiguate the proper answer from the distracter. 

Despite great efforts, existing methods cannot fully utilize these contexts. Specifically, embedding-based methods~\cite{TransE,RotatE} need expensive re-training to embed unseen entities; GNN-based methods~\cite{RGCN,GraIL} are less robust when few connections between existing and new entities are available; path-based methods~\cite{MINERVA,BERTRL} rely strongly on the existence and reliability of reasoning paths between certain entities; text-enhanced methods disregard entity type properties in KGs.  

A direct and promising approach to effectively utilize the three types of contexts is the integration of large language models~(LLMs). On the one hand, LLMs, trained on extensive corpora, possess a fundamental understanding of the type of KG entities. On the other hand, the strong semantic understanding and reasoning capabilities enable LLMs to capture crucial information from triples and paths~\cite{GenTKG}. \textit{Nevertheless, existing LLM-based KGC methods}~\cite{KICGPT,DIFT} \textit{can only re-rank candidate answers provided by previous KGC approaches}. Consequently, they are inevitably constrained by the limitations of preceding models.
\textit{Moreover, these approaches rely on additional triples from the validation set} for in-context demonstration~\cite{KICGPT} or supervised fine-tuning~\cite{DIFT}, \textit{which can lead to severe information leakage when being applied to inductive scenarios}. 

This paper proposes ``\textbf{CATS}'', a novel \textbf{\underline{C}}ontext-\textbf{\underline{A}}ware approach for the inductive KGC task based on latent \textbf{\underline{T}}ype constraints and \textbf{\underline{S}}ubgraph reasoning.  
Considering the semantic gap between natural language sentences and structural KG triples, CATS fine-tunes and guides LLMs to assess the existence of potential missing triples from two perspectives. First, the \textit{Type-Aware Reasoning}~(TAR) module evaluates whether the candidate entity conforms to the implicit type constrained by the relation. 
Since explicit type annotations are not prevalent for entities in non-encyclopedic KGs (e.g. Wordnet), we instead assess whether the candidate head/tail entity and other head/tail entities connected by the same relation belong to the same entity type. 
Then, the \textit{Subgraph Reasoning}~(SR) module proposes a degree-based filtering mechanism to select meaningful paths, and takes relevant neighboring facts of the head and tail entity into consideration. The superior long-context understanding capabilities of LLMs allow the SR module to comprehensively evaluate whether different paths and neighboring facts support the existence of the specific triple. Finally, we ensemble the inference results based on the scorings obtained from the two modules mentioned above.

We conduct extensive experiments on three widely used datasets: WN18RR, FB15k237, and NELL-995. The best variant of CATS significantly outperforms state-of-the-art approaches in $16$ out of $18$ transductive, inductive, and few-shot settings with an average absolute improvement of $7.2\%$ in MRR and $10.1\%$ in Hits@$1$. These results highlight the importance of incorporating the three types of contexts in KGs. Furthermore, ablation studies on various LLMs and configurations show that the effectiveness of the proposed method does not rely on internal knowledge and the scale of LLMs. Our contributions are summarized as follows:
\begin{itemize}[leftmargin=1em]
    \setlength\itemsep{0.1em}
    \item We propose CATS, the first LLM-based inductive KGC solution capable of handling unseen entities without any external knowledge or prior inference results. 
    
    \item We devise two novel triple evaluation mechanisms based on latent type constraints, as well as the reasoning paths and neighboring facts within the local subgraph. 
   
    \item We conduct extensive experiments to evaluate the effectiveness of CATS in different settings, and discuss the contribution of each component and the LLM in detail. 
\end{itemize}

\section{Related Works}
\paragraph{Embedding-based methods. } The majority of KGC methods rely on KG embeddings, such as TransE~\cite{TransE}, RotatE~\cite{RotatE}, and GIE~\cite{GIE}. These methods learn a set of low-dimensional embeddings for each entity and relation within the KG with certain geometric assumptions, which are inherently transductive. However, they require costly retraining to handle unseen entities~\cite{BERTRL}, limiting their adaptability to inductive scenarios. 

\paragraph{Graph neural network~(GNN)-based methods. } Graph neural networks (GNNs)~\cite{DBLP:conf/nips/SongZK23a,DBLP:conf/cikm/SongZK23} are popular in natural language processing for modeling relationships between entities~\cite{DBLP:conf/aaai/MaSHLZK23}. GNN-based methods such as CompGCN~\cite{CompGCN}, RGCN~\cite{RGCN}, and WGCN~\cite{WGCN} embed entities in a KG by iteratively aggregating features from the local neighborhood. 
Nevertheless, these approaches struggle to produce meaningful embeddings for newly emerged entities with few links to existing ones, and are not applicable to entirely new graphs~\cite{BERTRL}. To conform to the inductive setting, GraIL~\cite{GraIL}, and TACT~\cite{TACT} embed entities in the local subgraph with their distances to the head and tail entities of the query triple. However, such an embedding approach fails to distinguish different entities that share the same relative position. Consequently, it cannot perform well when the subgraph of query triple is large. RED-GNN~\cite{RED-GNN} and Adaprop~\cite{Adaprop} enhance the message-passing mechanisms through progressive and adaptive propagation. Still, these methods fail to address the suboptimal performance of GNNs on sparse graph structures.

\paragraph{Path-based methods. } Path-based methods aim to mine rules from the co-existences of certain reasoning paths connecting the head and tail entities in a triple and the relation between them~\cite{RuleN}. In particular, DeepPath~\cite{DeepPath} and MINERVA~\cite{MINERVA} exploit random-walk with reinforcement learning to generate reasoning paths, while BERTRL~\cite{BERTRL} and KRST~\cite{KRST} leverage breadth-first search. 
However, the existence and quality of reasoning paths between unseen entities is not ensured~\cite{APST}, which inevitably limits their generality. 

\paragraph{Text-enhanced methods. } In addition to the graph structure, the textual information provided in the KG also entails valuable semantic knowledge~\cite{SSET}. Recently, several KGC methods such as KG-BERT~\cite{kgbert}, BERTRL~\cite{BERTRL}, and KRST~\cite{KRST} employ PLMs to embed entities, relations, and reasoning paths with textual labels and descriptions. 
APST~\cite{APST} further introduces incomplete anchor paths for unseen entities that are not connected with any reasoning paths, achieving state-of-the-art performance. As the authors stated in~\cite{BERTRL}, combining multiple reasoning paths encourages knowledge interactions. Nevertheless, their BERT-based backbone PLMs~\cite{BERT} can only each reasoning path independently, leaving significant room for improvements. 

\section{Preliminaries} 

\begin{figure}
    \centering
    \includegraphics[width=0.45\textwidth]{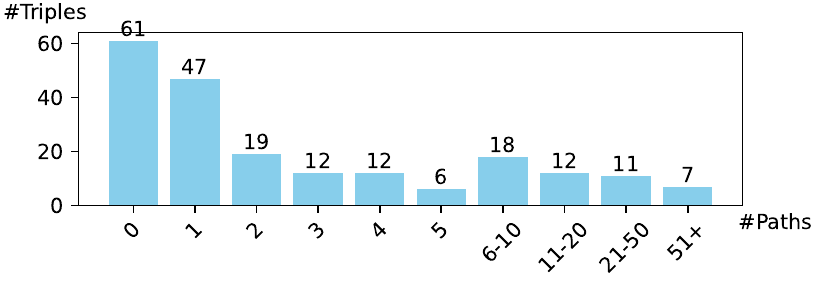}
    \caption{The statistics on the existence of reasoning paths for the query triples in the FB15k-237 (inductive) dataset. }
    \label{figure_stat}
\end{figure}

\paragraph{Problem specification. }A knowledge graph (or ``KG'' denoted as $\mathcal{G}=\{\mathcal{E}, \mathcal{R}, \mathcal{T}\}$ consists of a set of triples $\mathcal{T}=\{(h,r,t)\}$ where head and tail entities $h, t\in \mathcal{E}$, relation $r\in\mathcal{R}$. $\mathcal{E}$ and $\mathcal{R}$ denote the set of entities and relations, respectively. 
Following the convention of~\cite{GraIL,KRST,APST}, the task of inductive knowledge graph completion is defined as follows:

\paragraph{Definition 1 (Inductive KGC): } \textit{Given a training graph $\mathcal{G}_{\text{train}} \allowbreak= \{\mathcal{E}_{\text{train}}, \allowbreak\mathcal{R}_{\text{train}}, \mathcal{T}_{\text{train}}\}$ and a test graph $\mathcal{G}_{\text{test}} = \{\mathcal{E}_{\text{test}}, \allowbreak\mathcal{R}_{\text{test}}, \mathcal{T}_{\text{test}}\}$, the inductive KGC task aims to complete the missing head or tail entity from a set of query triples $Q=\{h_q, r_q, t_q\}_{q=1}^{|Q|}$ such that $\mathcal{E}_{\text{train}} \cap \mathcal{E}_{\text{test}} = \emptyset$, $\mathcal{R}_{\text{test}} \subseteq \mathcal{R}_{\text{train}}$, $\forall q, h_q, t_q \in \mathcal{E}_{\text{test}}, r_q \in \mathcal{R}_{\text{test}}$}.

The settings of the inductive KGC task ensure that entities in the training and test KGs form two disjoint sets. Only the triples in the training graph can be used in model training, while triples in the test graph are provided as evidence for query triple completion. Handling unseen entities requires the model to have inductive reasoning capabilities. 

\paragraph{Type properties of entities. } Apart from triples, entities within a  KG are often annotated with entity types (or categories) in an ontological taxonomy~\cite{JOIE}. For instance, in Freebase~\cite{Freebase}, entity ``\textit{Albert Einstein}'' belongs to the ``\textit{/scientist/physicist}'' type. In general, entity types provide a high-level summary of the salient properties of their instance entities, which play a crucial role in judging whether a specific entity is a plausible head or tail of a specific query relation. 
Nonetheless, explicit type annotations are typically scarce for entities in non-encyclopedic KGs like Wordnet. 

\paragraph{Reasoning paths. } Since entities in the test graph are not encountered during training, recent state-of-the-art methods~\cite{KRST,APST} leverage reasoning paths to make predictions. 
\paragraph{Definition 2 (Reasoning path): }\textit{Given a query triple $(h_q, r_q, t_q)$, a reasoning path is a sequence of triples that connects head entity $h_q$ and tail entity $t_q$. Formally, we have:}
\begin{equation}
    p(h_q, t_q) = h_q \xrightarrow{r_0} e_1, \xrightarrow{r_1} e_2, \xrightarrow{r_2}, ..., \xrightarrow{r_{n-1}} t_q,
\end{equation}
which satisfies $\forall i, (e_i, r_{i}, e_{i+1}) \in \mathcal{T}_{x}$ and $\forall j, r_j \neq r_q$. $\mathcal{T}_{x}$ denotes $\mathcal{T}_{\text{train}}$ during training and $\mathcal{T}_{\text{test}}$ for model evaluation. However, the existence and quality of reasoning paths are not guaranteed, especially in few-shot scenarios. For instance, our statistics in Figure~\ref{figure_stat} show that reasoning paths are unavailable for $61$ of the $205$ query triples in the test split of the FB15k-237 (inductive) dataset.  

\begin{figure*}
    \centering
    \includegraphics[width=0.95\textwidth]{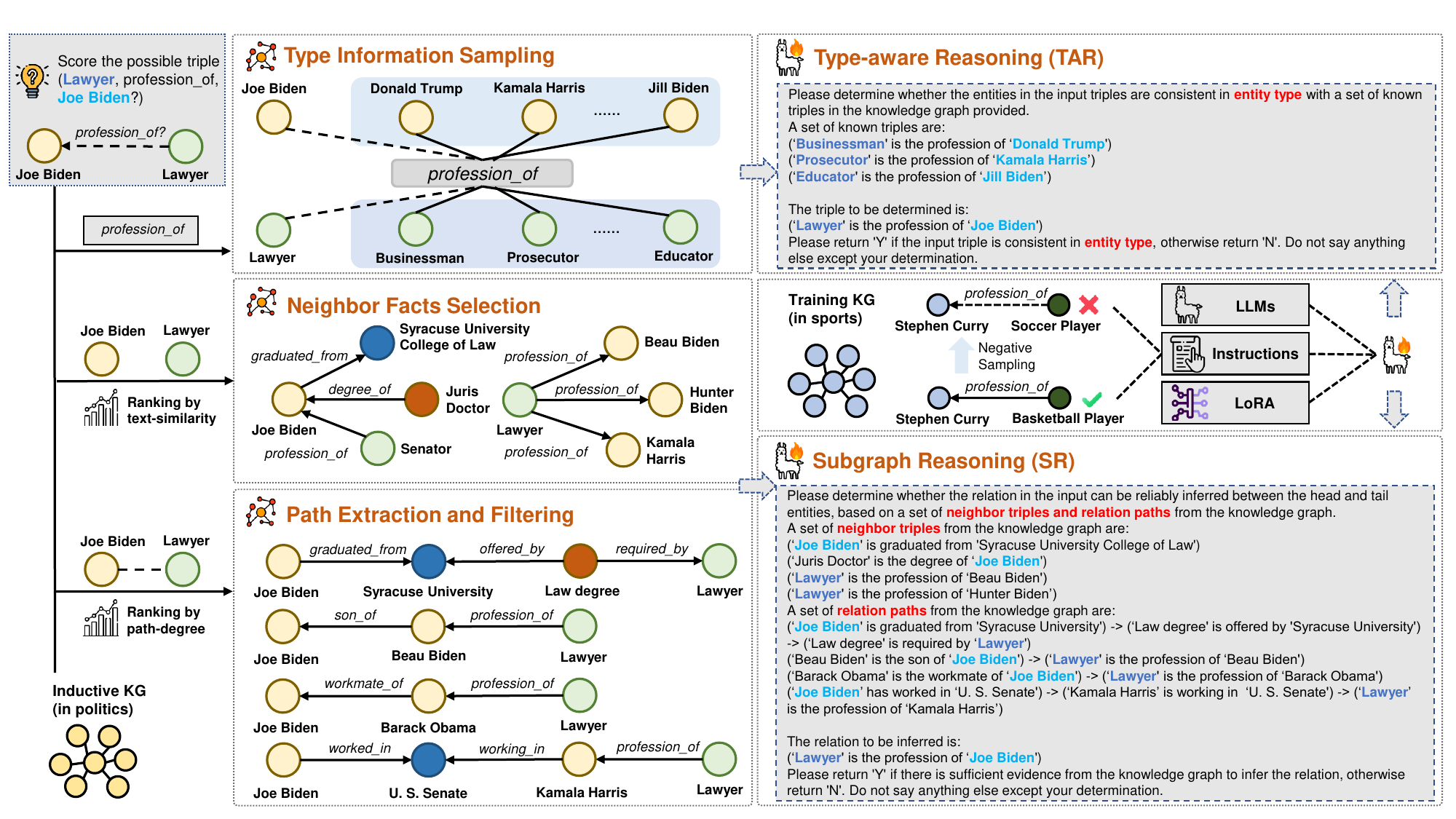}
    \caption{The end-to-end pipeline of the proposed CATS framework. }
    \label{fig_framework}
\end{figure*}

\section{Methodology}
Figure~\ref{fig_framework} shows the end-to-end architecture of the proposed CATS framework. To get rid of the reliance on explicit type annotations, we devise a more generalized approach to exploit latent type constraints w.r.t. relations in the Type-Aware Reasoning~(TAR) module. 
In addition, we incorporate relevant neighboring facts and reasoning paths to support triple assessment in the Subgraph Reasoning~(SR) module. 
Furthermore, we discuss our LLM supervised fine-tuning~(SFT) strategy, and aggregate our final predictions for query triples. 
\subsection{Type-aware Reasoning (TAR)}
In KGs, entities connected by the same relation often possess similar attributes and characteristics, thereby belonging to the same entity type. Therefore, when determining whether an unknown entity is the head or tail entity of a triple, we need to confirm that the entity conforms to the type dictated by the relation. In practice, type properties are not explicitly available for all entities in the KG. A direct solution is to employ LLMs to annotate type information for each candidate entity, and to summarize common type for entities connected by the particular relation. However, an entity may belong to multiple types with different granularities in different domains. For instance, in Freebase, entity \textit{Nick Mason} corresponds to type \textit{person}, \textit{film actor}, and \textit{book author}. Without clear guidance from explicit type annotations, the type output from the LLM will be unstable. 

Instead of explicitly conducting entity typing, we guide the LLM to evaluate the plausibility of a triple by implicitly considering the type relevance between the candidate head/tail entity and other head/tail entities connected by the same relation. 
Figure~\ref{fig_framework} (top-right) shows the detailed prompts. For each query triple $(h_q, r_q, t_q)$, we first sample a set of $k$ triples $\mathcal{S}_r$ with the same relation $r_q$ 
as demonstrations. Formally, we have $\mathcal{S}_r = \{(h, r_q, t)|h,t \in \mathcal{E}\setminus\{h_q, t_c\}\}$. Then, we linearize these structural triples with the textual labels of entities and relations, which allows the LLM to summarize a latent type triple e.g. (\textit{art work}, \textit{nominated for}, \textit{award}) between the set of head and tail entities. Finally, we provide the linearized query triple to the LLM, and ask the LLM to output ``Y'' if the query triple is consistent with the same pattern in terms of entity types and ``N'' otherwise. 

We fine-tune the LLM with a contrastive learning strategy. For each triple $(h, r, t)\in \mathcal{T}_{\text{train}}$, we construct negative samples by replacing the head or tail entity with a random entity from the training graph $\mathcal{G}_{\text{train}}$. Then, we utilize the following loss function for supervised fine-tuning:
\begin{align}
    \mathcal{L}_{\text{TAR}} =&-\sum_{(h,r,t)\in\mathcal{T}_{\text{train}}}\log{p(\hat{y}=\text{`Y'}|\mathcal{S}_r; \Theta)} \notag
    \\ &-\sum_{(h,r,t)\notin\mathcal{T}_{\text{train}}}\log{p(\hat{y}=\text{`N'}|\mathcal{S}_r; \Theta))}, 
\end{align}
where $\hat{y}$ denotes the first output token generated by the LLM, $\Theta$ denotes the model parameters, $p(\hat{y}=\text{`Y'})$ is the estimated probability that triple $(h,r,t)$ holds.

\subsection{Subgraph Reasoning (SR)}
Within KGs, the knowledge about an entity is manifested in its local subgraph~\cite{SSET}.
Simply considering the type of an entity is insufficient to assert that the candidate entity should be connected to certain entities with a specific relation. Recent studies~\cite{BERTRL,KRST} have shown that reasoning paths provide direct evidence for the existence of a particular relation between two entities. Nevertheless, limited by the power of BERT-like pre-trained language models, each reasoning path has to be independently encoded and considered, which may lead to unreliable relation prediction results. Inspired by the powerful reasoning capabilities of LLMs, we can model the interactions between multiple reasoning paths and neighboring facts of the two entities in a query triple. 

\paragraph{Path extraction and filtering. } Following the convention of~\cite{KRST,APST}, we leverage breadth-first search~(BFS) to extract reasoning paths connecting the two entities of the query triple. 
We only retain reasoning paths with a length less than or equal to $n$, as extraordinarily long paths contribute less to relation prediction. Moreover, we find that some high-frequency relations, such as ``\textit{has gender}'' and ``\textit{has color}'', are meaningless for assessing the existence of other relations~\cite{KRST}. On the contrary, infrequent fine-grained relations such as ``\textit{appear in film}'' usually offer more precise evidence. Hence, we devise a degree-based filtering mechanism to find out meaningful paths. Specifically, given a reasoning path $p(h_q, t_q)$, we count the occurrences $o_r$ for each relation $r \in p(h_q, t_q)$ in the training triple set $\mathcal{T}_{\text{train}}$. Then, we compute the degree of the reasoning path $d_{p(h_q,t_q)}$ by summing up the occurrences of all relations within the path. Formally, we have: 
\begin{equation}
    d_{p(h_q,t_q)} = \sum_{r \in p(h_q,t_q)} o_r = \sum_{r \in p(h_q,t_q)}\sum_{(h,r',t)\in \mathcal{T}_{\text{train}}} \mathds{1}(r=r'),
\end{equation}
where $\mathds{1}(\cdot)$ denotes the identifier function. Finally, for each query triple $(h_q, r_q, t_q)$, we select $\beta$ reasoning paths $\mathcal{P}_{(h_q, t_q)}$ with the lowest degrees $d_{p(h_q,t_q)}$ for assessment, while the others are filtered out. 

\paragraph{Neighboring facts selection. }
Since the existence of reasoning paths is not guaranteed, we further adopt neighboring facts of the head and tail entities of the query triple as supplementary contexts. Specifically, for each query triple $(h_q, r_q, t_q)$, we first collect supporting triples containing the head entity $h_q$ or the tail entity $t_q$ from the training graph.~\footnote{In inductive scenarios, supporting triples are selected from the test graph during the evaluation phase.} Then, we embed the query triple and each supporting triple with ``bge-small-en v1.5''~\cite{bgem3} sentence transformer. To safeguard the accuracy of our assessment against irrelevant neighboring information, we select top $\sigma$ supporting triples which the embeddings have the highest cosine similarities to the query triple.  Formally, we have 
\begin{equation}
    \mathcal{T}_{h_q} = \argmax_{(h_q, r, t)\in \mathcal{T}_{\text{train}}} \cos(f_{\text{bge}}(h_q,r,t), f_{\text{bge}}(h_q,r_q,t_q)),
\end{equation}
\begin{equation}
    \mathcal{T}_{t_q} = \argmax_{(h, r, t_q)\in \mathcal{T}_{\text{train}}} \cos(f_{\text{bge}}(h,r,t_q), f_{\text{bge}}(h_q,r_q,t_q)),
\end{equation}
where $cos(a,b)=\frac{a\cdot b}{{||a||}_2\cdot{||b||}_2}$, $\mathcal{T}_{h_q}$ and $\mathcal{T}_{t_q}$ are selected supporting triples for $h_q$ and $t_q$. Similarly, we design appropriate prompts to instruct the LLM to output ``Y'' if the given query triple can be supported by the aforementioned reasoning paths and neighboring triples and ``N'' otherwise (Figure~\ref{fig_framework} bottom). Then, we use the following loss function to fine-tune the LLM:
\begin{align}
    \mathcal{L}_{\text{SR}} =&-\sum_{(h,r,t)\in\mathcal{T}_{\text{train}}}\log{p(\hat{y}=\text{`Y'}|\mathcal{P}_{(h_q, t_q)}, \mathcal{T}_{h_q}, \mathcal{T}_{t_q}; \Theta)} \notag
    \\ &-\sum_{(h,r,t)\notin\mathcal{T}_{\text{train}}}\log{p(\hat{y}=\text{`N'}|\mathcal{P}_{(h_q, t_q)}, \mathcal{T}_{h_q}, \mathcal{T}_{t_q}; \Theta)}. 
\end{align}

\subsection{Triple Scoring}
During the inference stage, the commonly adopted evaluation metrics require the KGC model to rank (or score) each masked triple in the form of $(h,r,?)$ or $(?,r,t)$ with a set of candidate entities. 
Motivated by the complementary relationship between type properties and structural context, we compose the final scoring of a triple $s(h,r,t)$ by ensembling the probabilities that the LLM outputs ``Y'' based on the two kinds of prompts. Formally, we have
\begin{align}
    s(h,r,t) &= \frac{1}{2}\left(p(\hat{y}=\text{`Y'}|\mathcal{S}_r; \Theta) \right.\notag
    \\ &+ \left.p(\hat{y}=\text{`Y'}|\mathcal{P}_{(h, t)}, \mathcal{T}_{h}, \mathcal{T}_{t}; \Theta) \right).
\end{align}

\begin{table}[tbh!]
    \centering
    \footnotesize
    \begin{tabular}{lcccc}
        \toprule
        Dataset & Data splits & $|\mathcal{R}_G|$ & $|\mathcal{E}_G|$ & $|T_G|$ \\
        \midrule
        \midrule
        WN18RR & train & 9 & 2746 & 6670 \\
               & train-2000 & 9 & 1970 & 2002 \\
               & train-1000 & 9 & 1362 & 1001 \\
               & test-transductive & 7 & 962 & 638 \\
               & test-inductive & 8 & 922 & 1991 \\
        \midrule
        FB15k-237 & train & 180 & 1594 & 5223 \\
                  & train-2000 & 180 & 1280 & 2008 \\
                  & train-1000 & 180 & 923 & 1027 \\
                  & test-transductive & 102 & 550 & 492 \\
                  & test-inductive & 142 & 1093 & 2404 \\
        \midrule
        NELL-995 & train & 88 & 2564 & 10063 \\
                 & train-2000 & 88 & 1346 & 2011 \\
                 & train-1000 & 88 & 893 & 1020 \\
                 & test-transductive & 60 & 1936 & 968 \\
                 & test-inductive & 79 & 2086 & 6621 \\
        \bottomrule
    \end{tabular}
    \caption{Statistics of datasets.}
    \label{table_dataset_stats}
\end{table}

\begin{table*}[tbh!]
    \footnotesize
    \renewcommand\arraystretch{0.75}
    \centering
    \begin{tabular}{llcccccc}\toprule
          & & \multicolumn{3}{c}{Transductive} & \multicolumn{3}{c}{Inductive} \\
           \cmidrule(lr){3-5} \cmidrule(lr){6-8}
          Metric & Method & WN18RR & FB15k-237 & NELL-995 & WN18RR & FB15k-237 & NELL-995 \\ \midrule \midrule
    MRR   & RuleN & 0.669 & 0.674 & 0.736 & 0.780 & 0.462 & 0.710 \\
          & GRAIL & 0.676 & 0.597 & 0.727 & 0.799 & 0.469 & 0.675 \\
          & RED-GNN & 0.758 & 0.737 & 0.838 & 0.837 & 0.852 & 0.787 \\
          & Adaprop & 0.790 & 0.632 & 0.807 & 0.795 & 0.563 & 0.791 \\
          & MINERVA & 0.656 & 0.572 & 0.592 & - & - & - \\
          & TuckER & 0.646 & 0.682 & 0.800 & - & - & - \\
          & KG-BERT & - & - & - & 0.547 & 0.500 & 0.419 \\
          & BERTRL & 0.683 & 0.695 & 0.781 & 0.792 & 0.605 & 0.808 \\
          & KRST & 0.899 & 0.720 & 0.800 & 0.890 & 0.716 & 0.769 \\  
          & APST & 0.902 & 0.774 & 0.801 & 0.908 & 0.764 & 0.769 \\ \cmidrule(lr){2-8}
          & CATS (TAR) & 0.956 & 0.812 & 0.836 & 0.937 & 0.834 & 0.750 \\
          & CATS (SR) & \underline{0.972} & \underline{0.829} & \underline{0.869} & \textbf{0.992} & \underline{0.875} & \textbf{0.906} \\
          & CATS (full) & \textbf{0.978} & \textbf{0.843} & \textbf{0.885} & \underline{0.982} & \textbf{0.882} & \underline{0.861} \\ \midrule \midrule
    Hits@1 & RuleN & 0.646 & 0.603 & 0.636 & 0.745 & 0.415 & 0.638 \\
          & GRAIL & 0.644 & 0.494 & 0.615 & 0.769 & 0.390 & 0.554 \\
          & RED-GNN & 0.712 & 0.663 & 0.771 & 0.798 & 0.451 & 0.702 \\
          & Adaprop & 0.735 & 0.534 & 0.725 & 0.755 & 0.483 & 0.678 \\
          & MINERVA & 0.632 & 0.534 & 0.553 & - & - & - \\
          & TuckER & 0.600 & 0.615 & 0.729 & - & - & - \\
          & KG-BERT & - & - & - & 0.436 & 0.341 & 0.244 \\
          & BERTRL & 0.655 & 0.620 & 0.686 & 0.755 & 0.541 & 0.715 \\
          & KRST & 0.835 & 0.639 & 0.694 & 0.809 & 0.600 & 0.649 \\  
          & APST & 0.839 & 0.694 & 0.698 & 0.837 & 0.643 & 0.663 \\ \cmidrule(lr){2-8}
          & CATS (TAR) & 0.922 & 0.726 & 0.745 & 0.888 & 0.744 & 0.624 \\
          & CATS (SR) & \underline{0.951} & \underline{0.752} & \underline{0.792} & \textbf{0.984} & \underline{0.804} & \textbf{0.849} \\
          & CATS (full) & \textbf{0.962} & \textbf{0.776} & \textbf{0.820} & \underline{0.965} & \textbf{0.805} & \underline{0.783} \\
          \bottomrule
    \end{tabular}
    \caption{Transductive and inductive KGC results on WN18RR, FB15k-237 and NELL-995}
    \label{exp}
\end{table*}

\begin{table*}[tbh!]
\vspace{0.50cm}
\footnotesize
\renewcommand\arraystretch{0.75}
\centering
\begin{tabular}{llcccccccccccc}
            \toprule
          &         & \multicolumn{6}{c}{Transductive}                                                                    & \multicolumn{6}{c}{Inductive}                                                                       \\ \cmidrule(lr){3-8} \cmidrule(lr){9-14}
          & & \multicolumn{2}{c}{WN18RR}      & \multicolumn{2}{c}{FB15k-237}   & \multicolumn{2}{c}{NELL-995}    & \multicolumn{2}{c}{WN18RR}      & \multicolumn{2}{c}{FB15k-237}   & \multicolumn{2}{c}{NELL-995}    \\\cmidrule(lr){3-4} \cmidrule(lr){5-6} \cmidrule(lr){7-8} \cmidrule(lr){9-10} \cmidrule(lr){11-12} \cmidrule(lr){13-14}
          Metric & Method & 1000           & 2000           & 1000           & 2000           & 1000           & 2000           & 1000           & 2000           & 1000           & 2000           & 1000           & 2000           \\\midrule \midrule
    MRR   & RuleN   & 0.567          & 0.625          & 0.434          & 0.577          & 0.453          & 0.609          & 0.681          & 0.773          & 0.236          & 0.383          & 0.334          & 0.495          \\
          & GRAIL   & 0.588          & 0.673          & 0.375          & 0.453          & 0.292          & 0.436          & 0.652          & 0.799          & 0.380          & 0.432          & 0.458          & 0.462          \\
          & RED-GNN   & 0.144          & 0.301          & 0.250          & 0.519          & 0.296          & 0.469          & 0.818          & 0.826          & 0.482          & 0.503          & 0.692          & 0.737          \\
          & Adaprop   & 0.143          & 0.299          & 0.259          & 0.451          & 0.292          & 0.478          & 0.786          & 0.794          & 0.527          & 0.546          & 0.702          & 0.739          \\
          & MINERVA & 0.125          & 0.268          & 0.198          & 0.364          & 0.182          & 0.322          & -              & -              & -              & -              & -              & -              \\
          & TuckER  & 0.258          & 0.448          & 0.457          & 0.601          & 0.436          & 0.577          & -              & -              & -              & -              & -              & -              \\
          & KG-BERT & -              & -              & -              & -              & -              & -              & 0.471          & 0.525          & 0.431          & 0.460          & 0.406          & 0.406          \\
          & BERTRL  & 0.662          & 0.673          & 0.618          & 0.667          & 0.648          & 0.693          & 0.765          & 0.777          & 0.526          & 0.565          & 0.736          & 0.744          \\
          & KRST    & 0.871          & 0.882 & 0.696          & 0.701          & 0.743          & \textbf{0.781} & 0.886          & 0.878          & 0.679          & 0.680          & 0.745          & 0.738          \\ & APST    & 0.874 & 0.880 & 0.724 & 0.753 & \textbf{0.745} & \underline{0.767}          & 0.894 & 0.879 & 0.697 & 0.747 & 0.765 & 0.747 
          \\\cmidrule(lr){2-14}
          & CATS (TAR)    & \underline{0.925} & \underline{0.936} & \underline{0.774} & \underline{0.800} & 0.737 & 0.751          & 0.869 & 0.889 & 0.796 & 0.813 & 0.697 & 0.712 \\
          & CATS (SR)    & 0.879 & 0.926 & 0.734 & 0.780 & 0.680 & 0.679          & 0.898 & \textbf{0.956} & \underline{0.847} & \underline{0.876} & \textbf{0.854} & \textbf{0.871} \\
          & CATS (full)   & \textbf{0.932} & \textbf{0.952} & \textbf{0.787} & \textbf{0.824} & \underline{0.741} & 0.762          & \textbf{0.922} & \underline{0.953} & \textbf{0.862} & \textbf{0.877} & \underline{0.808} & \underline{0.829} \\  
    \midrule \midrule
    Hits@1 & RuleN   & 0.548          & 0.605          & 0.374          & 0.508          & 0.365          & 0.501          & 0.649          & 0.737          & 0.207          & 0.344          & 0.282          & 0.418          \\
          & GRAIL   & 0.489          & 0.633          & 0.267          & 0.352          & 0.198          & 0.342          & 0.516          & 0.769          & 0.273          & 0.351          & 0.295          & 0.298          \\
          & RED-GNN   & 0.108          & 0.267          & 0.196          & 0.449          & 0.214          & 0.379          & 0.777          & 0.785          & 0.380          & 0.422          & 0.549          & 0.605          \\
          & Adaprop   & 0.107          & 0.260          & 0.188          & 0.366          & 0.218          & 0.386          & 0.741          & 0.749          & 0.425          & 0.451          & 0.580          & 0.630          \\
          & MINERVA & 0.106          & 0.248          & 0.170          & 0.324          & 0.152          & 0.284          & -              & -              & -              & -              & -              & -              \\
          & TuckER  & 0.230          & 0.415          & 0.407          & 0.529          & 0.392          & 0.520          & -              & -              & -              & -              & -              & -              \\
          & KG-BERT & -              & -              & -              & -              & -              & -              & 0.364          & 0.404          & 0.288          & 0.317          & 0.236          & 0.236          \\
          & BERTRL  & 0.621          & 0.637          & 0.517          & 0.583          & 0.526          & 0.582          & 0.713          & 0.731          & 0.441          & 0.493          & 0.622          & 0.628          \\
          & KRST    & 0.790          & 0.810          & 0.611          & 0.602          & 0.628          & \textbf{0.678} & 0.811          & 0.793          & 0.537          & 0.524          & 0.637          & 0.629 \\
          & APST    & 0.798 & 0.813 & 0.632 & 0.665 & \underline{0.640} & 0.663          & 0.822 & 0.798 & 0.561 & 0.627 & 0.654 & 0.637 
          \\\cmidrule(lr){2-14}
          & CATS (TAR)   & \underline{0.871} & \underline{0.888} & \underline{0.678} & \underline{0.714} & 0.628 & 0.636          & 0.774 & 0.811 & 0.680 & 0.707 & 0.584 & 0.596 \\
          & CATS (SR)    & 0.802 & 0.874 & 0.631 & 0.681 & 0.574 & 0.563          & \underline{0.824} & \underline{0.915} & \underline{0.756} & \underline{0.800} & \textbf{0.767} & \textbf{0.790} \\
          & CATS (full)    & \textbf{0.887} & \textbf{0.918} & \textbf{0.702} & \textbf{0.750} & \textbf{0.648} & \underline{0.664}          & \textbf{0.864} & \textbf{0.923} & \textbf{0.776} & \textbf{0.802} & \underline{0.713} & \underline{0.746} \\\bottomrule
\end{tabular}
\caption{Few-shot KGC results on WN18RR, FB15k-237 and NELL-995}
\label{table_few_shot}
\end{table*}

\section{Experiments}
\subsection{Datasets and Evaluation Metrics}
We evaluate our proposed method on three widely adopted benchmark KGs WN18RR~\cite{WordNet}, FB15k-237~\cite{Freebase}, and NELL-995~\cite{NELL} with their transductive and inductive subsets. Following the convention of~\cite{BERTRL,KRST,APST}, we evaluate each query triple with $1$ ground truth entity and $49$ negative candidate entities. 
For a fair comparison, we use the same dataset splits and negative triples provided by~\cite{BERTRL}. 
The detailed statistics of datasets are available in Table~\ref{table_dataset_stats}. We score the plausibility of the query triple with each candidate entity and rank them in descending order. The performance of our model is evaluated based on two metrics: Mean Reciprocal Rank~(MRR) and Hits@$1$. 

\subsection{Baselines and Experiment Settings}
We compare CATS with embedding-based methods RuleN and TuckER, GNN-based method GraIL, Red-GNN and Adaprop, text-based methods KG-BERT, and path-based methods MINERVA, BERTRL, KRST and the state-of-the-art method APST. We select Qwen2-7B-Instruct~\cite{Qwen2-7B} as our backbone LLM. The experimental results on Llama-3-8B~\cite{llama3} and Qwen2-1.5B are also included in our ablation studies.
We employ LoRA~\cite{lora} for parameter-efficient fine-tuning, setting the rank to 16 and $\alpha$-value to 32. 
We use the AdamW optimizer~\cite{adamW} to fine-tune the LLM with a learning rate of $1\text{e-}4$, a per-device batch size of $2$, and a gradient accumulation step of $4$ iterations for $1$ epoch only. For each query triple, we sample $k=3$ supporting triples with the same relation, $\delta=6$ reasoning paths and $\sigma=6$ neighboring facts. We construct the instruction set by generating 12 negative samples for each positive triple in $\mathcal{T}_{\text{train}}$. All experiments are conducted on a server with 2 Intel Xeon Platinum 8358 processors and 8 NVIDIA A100 40G GPUs.~\footnote{Only 2 GPUs are used in our experiments. } On average, CATS takes $2.4$ hours for SFT and $1.43$s to evaluate and rank a single test sample.

\subsection{Main Results}
We evaluate the proposed CATS framework on the three datasets in transductive and inductive settings. The ``TAR'' and ``SR'' variants of CATS severally utilize the probabilities generated by corresponding modules to score and rank each candidate entity. Experimental Results in Table~\ref{exp} demonstrate that the CATS~(full) significantly and consistently outperforms all baseline methods. 
Most notably, CATS~(full) achieves \textbf{\textit{absolute Hits@1 improvements}} of $12.8\%$, $16.2\%$, and $6.8\%$ on the WN18RR, FB15k-237, and NELL-995 datasets under an \textbf{\textit{inductive setting}}. Correspondingly, the improvements in \textbf{\textit{transductive}} scenarios, namely $12.3\%$, $8.2\%$, and $9.1\%$, are also remarkable. 

Among all CATS' variants, the majority of best results (4 out of 6 cases) are achieved by the ``full'' variant, which exhibits the importance of considering latent type constraints and subgraph contexts. In comparison, the improvements observed with the TAR variant are less pronounced, indicating that while matching entity types may help filter out irrelevant entities, it is insufficient for delivering accurate relation predictions. Moreover, the TAR variant demonstrates improved effectiveness in transductive settings. The better performance can be credited to SFT, which enhances the LLM's understanding of the type properties of known entities. However, the training graph does not contain any contexts for unseen entities. Therefore, the LLM may not be able to precisely infer types for some of them, thereby compromising the performance of TAR in inductive scenarios. Conversely, the SR variant benefits from neighboring facts and reasoning paths sampled from the test graph, providing relevant contextual information for unseen entities and allowing it to achieve state-of-the-art performance on WN18RR and NELL-995 datasets in inductive settings. 

\subsection{Ablation Studies}
We examine the effectiveness of each component of the proposed CATS framework in different settings by answering the following research questions (RQs). 

\paragraph{RQ1: Can CATS generate plausible inference results in few-shot scenarios?} Table~\ref{table_few_shot} shows the experiment results on the three datasets with $1000$ and $2000$ triples in the training graph. In general, CATS achieves significant improvements for $10$ out of $12$ cases in terms of MRR compared to state-of-the-art methods.
For the remaining cases, the performance gap to the best baseline method is marginal. Considering the three variants of CATS, TAR outperforms SR in transductive settings, while SR performs better in inductive scenarios. The disparity in performance can be attributed to the following reasons: Reducing the number of triples in the training graph significantly decreases the average degree of each entity. Consequently, it becomes challenging to sample a sufficient number of neighboring facts and reasoning paths for entities in the query triple, which diminishes the effectiveness of the SR variant. However, in most KGs, the number of relations is considerably smaller than the number of entities ($|\mathcal{R}| << |\mathcal{E}|$). Hence, we are still able to retrieve sufficient triples with the same relation from the training graph, which sustains the desirable performance of the TAR variant, and ensures the robustness of the the proposed framework. In inductive scenarios, neighboring facts and reasoning paths are sampled from the supplementary test graph. Hence, reducing training triples has subtle negative impacts on the effectiveness of the SR variant. 

One may notice that CATS does not achieve stat-of-the-art performance on the NELL-995 dataset in transductive settings. This is attributed to the higher number of triples in the NELL-995 training graph. Selecting the subset of $1000$ or $2000$ triples results in an extremely sparse graph structure. Without external knowledge, the training graph may fail to provide sufficient support for the inference of certain triples, thereby lowering the performance ceiling. For the same reason, GNN-based methods such as RED-GNN~\cite{RED-GNN} and Adaprop~\cite{Adaprop} exhibit significant performance drops in few-shot scenarios. Moreover, KRST~\cite{KRST} and APST~\cite{APST} take advantage of extra knowledge from entity descriptions, allowing them to catch up with CATS.

\paragraph{RQ2: Do reasoning paths and neighboring facts improve the inference performance? } From the experimental results in Table~\ref{table_SR}, we observe that both reasoning paths and neighboring facts play a crucial role in enhancing inference performance. However, the contribution of neighboring triples is more pronounced. This reconfirms the key shortcoming of path-based methods, which struggle to assess query triples without suitable paths. 
In comparison, neighboring triples guarantee CATS's robustness and generality. Moreover, the performance decline resulting from the removal of the path filtering step emphasizes the effectiveness of the proposed degree-based filtering mechanism, further reaffirming that irrelevant reasoning paths may misdirect the evaluation of query triples. 

\begin{table*}[tbh!]
    \centering
    \footnotesize
    \begin{tabular}{llccccccc}
        \toprule
        & & \multicolumn{3}{c}{Transductive} & \multicolumn{3}{c}{Inductive} \\
        \cmidrule(lr){3-5} \cmidrule(lr){6-8}
        Metric & Configuration & WN18RR         & RB15k-237      & NELL-995       & WN18RR         & RB15k-237      & NELL-995       \\ 
        \midrule \midrule
    MRR & CATS (SR) & \textbf{0.972} & \textbf{0.829} & \textbf{0.869}  & \textbf{0.992} & \textbf{0.875} & \textbf{0.906} \\
        & - w/ NF only         & 0.962 & 0.824 & 0.861 & 0.971 & 0.873 & 0.895 \\
        & - w/ RP (filt.) only & 0.960 & 0.800 & 0.773 & 0.968 & 0.851 & 0.829 \\
        & - w/ RP only         & 0.954 & 0.801 & 0.769 & 0.965 & 0.844 & 0.821 \\
        \bottomrule \midrule
    Hits@1 & CATS (SR) & \textbf{0.951} & \textbf{0.752} & \textbf{0.792} & \textbf{0.984} & \textbf{0.804} & \textbf{0.849} \\
        & - w/ NF only         & 0.932 & 0.747 & 0.776 & 0.944 & 0.793 & 0.830 \\
        & - w/ RP (filt.) only & 0.932 & 0.714 & 0.666 & 0.941 & 0.768 & 0.729 \\
        & - w/ RP only         & 0.929 & 0.714 & 0.660 & 0.937 & 0.756 & 0.724 \\
        \bottomrule
    \end{tabular}
    \caption{Transductive and Inductive KGC performance (in MRR) on the SR variant with different structural contexts. Here NF denotes neighboring facts, RP denotes reasoning paths, and (filt.) refers to the application of path filtering.}
    \label{table_SR}
\end{table*}

\begin{table*}[tbh!]
    \centering
    \footnotesize
    \begin{tabular}{llccccccc}
        \toprule
        & & \multicolumn{3}{c}{Transductive} & \multicolumn{3}{c}{Inductive} \\
        \cmidrule(lr){3-5} \cmidrule(lr){6-8}
        Metric & LLM \& Config. & WN18RR & RB15k-237 & NELL-995 & WN18RR & RB15k-237 & NELL-995 \\
        \midrule \midrule
    MRR & Previous SOTA & 0.902 & 0.774 & 0.801 & 0.908 & 0.764 & 0.808 \\
        \cmidrule(lr){2-8}
        & Llama-3-8B (full) & 0.965 & 0.802 & 0.867 & 0.956 & 0.862 & 0.837 \\
        & Qwen2-1.5B (full) & 0.966 & 0.804 & 0.877 & 0.948 & 0.814 & 0.835\\
        & Qwen2-7B (comb.) & 0.967 & 0.830 & 0.861 & \textbf{0.985} & 0.859 & \textbf{0.885} \\
        & Qwen2-7B (full) & \textbf{0.978} & \textbf{0.843} & \textbf{0.885} & 0.982 & \textbf{0.882} & 0.861 \\
        & - w/o. TAR \& SR & 0.947 & 0.800 & 0.811 & 0.901 & 0.814 & 0.710 \\
        & - w/o. SFT & 0.225 & 0.197 & 0.177 & 0.207 & 0.179 & 0.208 \\
        & - w/o. all & 0.199 & 0.140 & 0.156 & 0.174 & 0.144 & 0.138 \\
        \bottomrule \midrule
    Hits@1 & Previous SOTA  & 0.839 & 0.694 & 0.698 & 0.837 & 0.643 & 0.715 \\
        \cmidrule(lr){2-8}
        & Llama-3-8B (full) & 0.942 & 0.718 & 0.794 & 0.931 & 0.783 & 0.750 \\
        & Qwen2-1.5B (full) & 0.940 & 0.720 & 0.809 & 0.912 & 0.722 & 0.748 \\
        & Qwen2-7B (comb.)  & 0.941 & 0.750 & 0.779 & \textbf{0.971} & 0.778 & \textbf{0.811} \\
        & Qwen2-7B (full)   & \textbf{0.962} & \textbf{0.776} & \textbf{0.820} & 0.965 & \textbf{0.805} & 0.783 \\
        & - w/o. TAR \& SR  & 0.905 & 0.705 & 0.704 & 0.824 & 0.707 & 0.572 \\
        & - w/o. SFT        & 0.132 & 0.103 & 0.081 & 0.101 & 0.093 & 0.009 \\
        & - w/o. all        & 0.152 & 0.093 & 0.103 & 0.128 & 0.093 & 0.087 \\
        \bottomrule
    \end{tabular}
    \caption{Transductive and Inductive KGC performance (in MRR) with different LLMs and configurations. 
    }
    \label{table_config}
\end{table*}

\paragraph{RQ3: Can we simply attribute the performance improvement to extra knowledge inherent in the LLM?} We conduct additional experiments by presenting the query triple and corresponding entities to the LLM, prompting the model to make judgments with its internal knowledge. However, experimental results in Table~\ref{table_config} indicate that the LLM fails to make reliable assessments on KG triples in such a zero-shot setting (see Qwen2-7B w/o. all), showing that CATS does not benefit from the LLM's internal knowledge. Furthermore, the pre-trained LLM cannot adequately comprehend the contextual information (e.g., paths and triples) outlined in our prompts without proper guidance (see Qwen2-7B w/o SFT). The unsatisfactory performance underscores the substantial semantic gap between natural language sentences and KG triples, emphasizing the importance of SFT. 

\paragraph{RQ4: Can we simply attribute the performance improvements to the power of LLMs?} 
We conduct an extra experiment by directly fine-tuning the LLM to evaluate the plausibility of query triples based solely on the triple itself (see Table~\ref{table_config} w/o. TAR \& SR). The significant performance decline indicates the following: despite possessing enhanced semantic understanding capabilities, the LLM does not inherently know the appropriate method to evaluate a triple. On the one hand, the backbone LLM in this variant cannot recognize the importance of type relevance between the target entity and entities connected by the same relation. 
On the other hand, the process of SFT is insufficient to inject knowledge stored in KGs into the LLM. This reaffirms the significance of providing relevant guidance and structural contexts in the KGC task. Furthermore, we investigate whether combining prompts from the two reasoning modules improves the model performance (see Table~\ref{table_config} (comb.)). Our experiments show that the current adopted separated setting achieves better results in most of the ($4$ out of $6$) cases.

\paragraph{RQ5: How does the selection of backbone LLM affect the experimental results?}
In Table~\ref{table_config}, we evaluate the performance of the proposed CATS framework on three LLMs, namely Llama3-8B, Qwen2-1.5B, and Qwen2-7B. The experimental results show that CATS significantly and consistently surpasses the state-of-the-art method with all these LLMs, demonstrating its broad effectiveness across different model scales and architectures. 
Most notably, CATS can still achieve desirable results with an $1.5$B model, which significantly reduces the average inference time from $1.43$s to $0.51$s, showcasing a perfect balance between performance and efficiency. Furthermore, the performance improvements observed with the Qwen2-7B model indicate that increasing the model size is likely to yield better results. Since comparison among different LLMs is not the primary focus, this paper does not explore the performance of CATS with larger LLMs due to time and resource constraints. 

\section{Conclusion}
In this paper, we propose CATS, a novel context-aware approach for knowledge graph completion. CATS is designed to guide the LLM to assess the plausibility of query triples based on latent type constraints, selected reasoning paths, and relevant neighboring facts. With sufficient guidance from proper prompts and SFT, CATS achieves state-of-the-art performance in transductive, inductive, and few-shot scenarios, showing its robustness and generality. 
Overall, CATS demonstrates the potential of leveraging LLMs in conducting knowledge-intensive reasoning tasks on structural data. In the future, we aim to incorporate LLMs in more complicated KG-related tasks such as complex query answering.

\section*{Acknowledgment}
The work described in this paper was partially supported by the Research Grants Council of the Hong Kong Special Administrative Region, China (CUHK 14222922, RGC GRF 2151185).
\bibliography{references}

\end{document}